\documentclass[10pt,twocolumn,letterpaper]{article}

\usepackage{wacv}
\usepackage{times}
\usepackage{epsfig}
\usepackage{graphicx}
\usepackage{amsmath}
\usepackage{amssymb}

\usepackage[sort,nocompress]{cite}

\usepackage{microtype}
\usepackage{subfigure}
\usepackage{booktabs}
\usepackage[ruled,vlined]{algorithm2e}

\usepackage{dsfont}

\usepackage{multirow}

\def\wacvPaperID{80} 

\wacvfinalcopy 

\ifwacvfinal
\def\assignedStartPage{1} 
\fi

\ifwacvfinal
\usepackage[breaklinks=true,bookmarks=false]{hyperref}
\else
\usepackage[pagebackref=true,breaklinks=true,colorlinks,bookmarks=false]{hyperref}
\fi

\ifwacvfinal
\else
\fi

\graphicspath{{img/background/}}

\newcommand{\approptoinn}[2]{\mathrel{\vcenter{
			\offinterlineskip\halign{\hfil$##$\cr
				#1\propto\cr\noalign{\kern2pt}#1\sim\cr\noalign{\kern-2pt}}}}}

\begin{document}

\title{Simple and Effective Balance of Contrastive Losses}

\author{Arnaud Sors$^\dagger$ \quad Rafael Sampaio de Rezende$^\dagger$ \quad Sarah Ibrahimi$^\ddag$\thanks{Work partially done while Sarah was interning at NAVER LABS Europe.} \quad Jean-Marc Andreoli$^\dagger$ \\
$\dagger$ NAVER LABS Europe \quad $\ddag$ University of Amsterdam
}

\maketitle

\begin{abstract}
    Contrastive losses have long been a key ingredient of deep metric learning and are now becoming more popular due to the success of self-supervised learning.
    Recent research has shown the benefit of decomposing such losses into two sub-losses which act in a complementary way when learning the representation network:
    a positive term and an entropy term.
    Although the overall loss is thus defined as a combination of two terms, the balance of these two terms is often hidden behind implementation details
    and is largely ignored and sub-optimal in practice.
    In this work, we approach the balance of contrastive losses as a hyper-parameter optimization problem, and propose a coordinate descent-based search method that efficiently find the hyper-parameters that optimize evaluation performance.
    In the process, we extend existing balance analyses to the contrastive margin loss, include batch size in the balance, and explain how to aggregate loss elements from the batch to maintain near-optimal performance over a larger range of batch sizes.
    Extensive experiments with benchmarks from deep metric learning and self-supervised learning show that optimal hyper-parameters are found faster
    with our method than with other common search methods.
\end{abstract}

\section{Introduction}
\label{sec:intro}

\begin{figure*}
    \centering
    \small
    \def\svgwidth{\linewidth}
    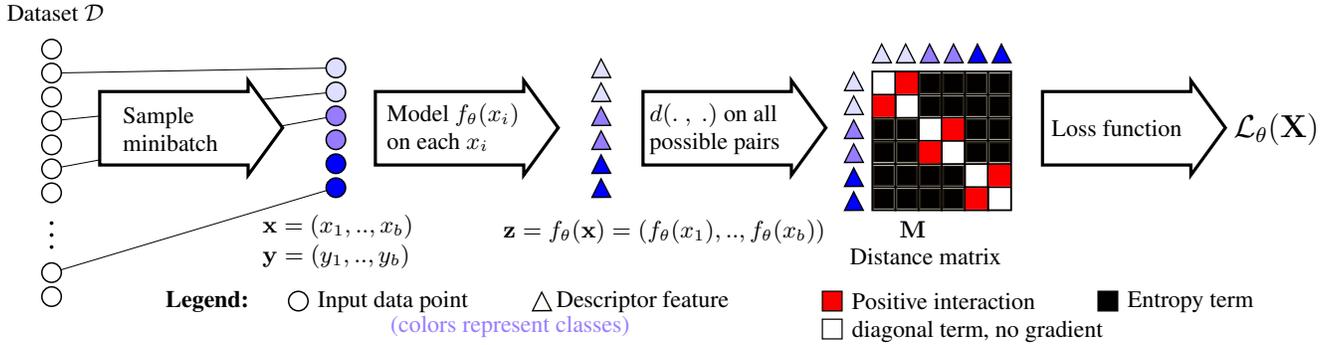
    \caption{An overview of the training pipeline}
    \label{fig:pipeline}
\end{figure*}

A neural network for deep metric learning (DML) maps data to an embedding space in which similar data points are close together and dissimilar data points are further apart, so that a simple non-parametric distance function applied on top of output embeddings behaves as a rich distance metric in data space. DML plays a major role in image and text retrieval~\cite{gordo2016deep,jing2015visual,ding2020rocketqa}, cross-modal retrieval~\cite{wang2016learning}, and person re-identification~\cite{yi2014deep,schroff2015facenet,hermans2017defense}. Recently the learning setup of DML has been successfully used for contrastive self-supervised learning (SSL)~\cite{chen2020simple}, in which augmentations of a sample are created to serve as positives and replace instance labels. SSL has proven extremely successful and is now established as one of the leading methods for learning general-purpose image representations~\cite{goyal2021self,caron2020unsupervised,chen2020simple,he2020momentum,grill2020bootstrap,caron2020emerging,assran2020semi}.

SSL and DML are very similar from an optimization point of view. In both cases, the evaluation metric is a non-differentiable measure, which is why training needs to rely on a surrogate loss function. There are many such loss functions, but recent experimental benchmarks have shown that this field lacks fair comparisons, and simple pair-based contrastive losses perform as well as more elaborate ones if correctly tuned~\cite{musgrave2020metric,roth2020revisiting,fehervari2019unbiased}.
Such losses distinguish between similar/positive samples and dissimilar/negative samples in a batch during training. However, even these most simple losses are often considered difficult to optimize, and practitioners often rely on various training methods which can be useful in practice but interact in an unclear way with the optimization process: for example negative mining~\cite{schroff2015facenet,wu2017sampling, stylianou2020a, stylianou2020b},
memory banks~\cite{wu2018unsupervised,wang2020cross,xiao2017joint,he2020momentum,tian2019contrastive,misra2020self}, and positive-only formulations~\cite{grill2020bootstrap,tian2020understanding,chen2020exploring}.

In this paper we go back to basics and focus on the following research problem for pair-based contrastive losses: how should their positive and entropy parts be balanced in order to maximize model performance?
Although this balance is simple and essential, it is often ignored because it is not explicitly coded as a hyper-parameter.
We ask ourselves which hyperparameters should be included in the balance and how to easily find their best setting.

We build on the work by Wang and Isola~\cite{wang2020understanding}, who studied the specific case of SSL with the InfoNCE loss. Here we consider both DML and SSL, extend their analysis to the contrastive margin loss, and include batch size in the analysis~\footnote{\cite{wang2020understanding} also did, but only linearly coupled with the learning rate which is one less degree of freedom. }.
We measure experimentally the influence of the balance on evaluation performance, and observe that tuning it is crucial to good evaluation performance. We also explore how to do this efficiently and provide a simple hyper-parameter optimization method for that purpose. More precisely, we make the following contributions:
\begin{itemize}
\item \textbf{Balance}. Extending Wang and Isola~\cite{wang2020understanding} to the contrastive margin loss and to one more degree of freedom, we decompose both losses into a \textit{positive} and an \textit{entropy} term, and study in detail how the overall balance influences the transfer-task performance (mAP, R-mAP, or accuracy).
      We observe that tuning this balance results in improvements up to 9.8$\%$ in test performance.
      For the contrastive margin loss we show how a proper choice of batch aggregation makes it possible to maintain near-maximal performance over a larger range of batch sizes.

\item \textbf{HPO}. We provide a method for efficiently finding the optimal hyper-parameter setting. This method consists of a coordinate descent in a reparameterized space, and accelerates hyper-parameter search, yielding an improvement of four points in AUC@20.
\end{itemize}

\section{Related work}
\label{sec:related}

\textbf{Loss functions}. Training losses for DML and contrastive SSL can be used interchangeably.
A plethora of such loss functions exist.
They can be sub-categorized based on whether they involve single points~\cite{qian2019softtriple}, pairs~\cite{hadsell2006dimensionality,oord2018representation},
triplets~\cite{chechik2010large}, or lists~\cite{liu2011learning,cao2007learning} of datapoints $x_i$.
In spite of this diversity, recent experimental studies~\cite{musgrave2020metric,roth2020revisiting} seem to show that
simple well-tuned pair-based contrastive losses are hard to beat when they are properly tuned.

\textbf{Decomposition of contrastive losses}. Wang (Xun) \etal~\cite{wang2019multi} decompose a number of pair-based losses into a positive and negative term.
Out of this understanding they develop the Multi-Similarity loss which is a relatively complex combination of mining and weighing.
Closer to our work, Wang (Tongzhou) and Isola~\cite{wang2020understanding} recently showed that the infoNCE loss
can be viewed as a two-task problem - a term on positives called \textit{alignment} term and a term enforcing
\textit{uniformity} of representations, which we call here the \textit{entropy} term.
They also show in the case of SSL the interest of 1D-balancing these two terms.
In this work we extend this understanding to the contrastive margin loss which is different from a batch aggregation point of view, and to image retrieval.
We consider two or three parameters in the balance rather than one.
Also, on top of observing that a correct balance helps, we also study how to find it efficiently,
and propose a hyper-parameterization method for to do so.
Finally, Zhang \etal~\cite{zhang2021rethinking} also tune the balance between positive and negative terms in a contrastive loss,
but from a different perspective: by changing the shape of the function which associates a similarity value to a loss value.
This can be viewed as a kind of in-batch mining, which changes the mean gradient per term as a side effect.
This approach is similar to ours in purpose, but more complicated as it acts on the shape of the loss function as a secondary means to control its average value.

\textbf{Training practices.} It used to be common practice in DML to \textit{select} interactions to include in the loss,
either ahead of time by performing out-of-batch negative mining~\cite{schroff2015facenet,gordo2017end,harwood2017smart}, or from within the minibatch~\cite{wu2017sampling}.
Another common practice was to use each data sample in a single pair only~\cite{radenovic2018fine}.
However, it was subsequently discovered that no mining and the use of all possible interactions~\cite{oh2016deep,sohn2016improved} is simpler and usually works best.
This is now the standard practice.
Another, more recent, practice in SSL and DML is the use of a memory bank~\cite{wang2020cross,he2020momentum}.
As for negative mining, while such memory banks have been shown useful in practice, our work questions how much of their benefits
result from the fact that they modify the balance of the loss.

\section{Pair-based contrastive losses}
\label{sec:background}

We now describe the pipeline shown in Figure~\ref{fig:pipeline} used for contrastive training in DML and SSL
and the associated loss functions.

\subsection{Pipeline}
\label{subsec:pipeline}

\begin{figure*}
    \centering
    \includegraphics{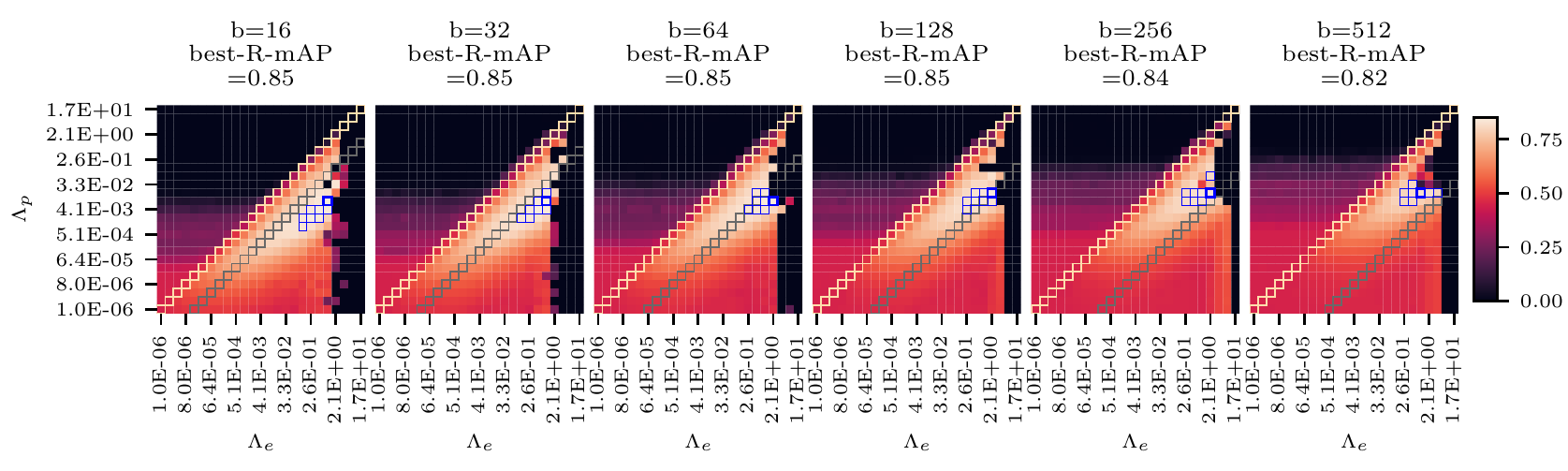}
    \caption{A visualization of image retrieval test performance of a network trained with a contrastive margin loss
        on Omniglot, as a function of $h=(\Lambda_p, \Lambda_e, b)$ (log-space visualization). Color represents test performance.
        Each square subfigure is a slice of constant batch size. Grey (resp. beige) diagonals correspond to the hyperplane of
        configurations reachable using a \textit{global} (resp. \textit{separate}) average as described in~\ref{ssec:ctrm_norm}, i.e. \textit{without balancing}.
        Blue squares are the top-8 configurations for each batch size.
        \textbf{There is only one batch size (128) for which the best configuration is reachable without balance.   }}
    \label{fig:hp_omniglot}
    \end{figure*}

We assume a training dataset $\cal D$ where each item contains an instance $x_n$ and a label $y_n$, which is implicit in the case of self-supervised learning.
The first step in training is the sampling of a training minibatch of size $b$. 
We obtain a minibatch $(\mathbf x, \mathbf y)$ where labels $\mathbf y$ are arranged by pairs as shown on Figure~\ref{fig:pipeline} (the discussion also easily extends to groups of more than two).
In DML labels $\mathbf y \in [\![1, C]\!]^b$ come from class labels, whereas in SSL labels $\mathbf y \in [\![1, \frac{b}{2}]\!]^b$ are artificial and come from data augmentations~\cite{chen2020simple}.
In the second step, minibatch embeddings $z_i=f_\theta(x_i)$ are obtained by applying an embedding function $f_\theta$ parameterized by $\theta$ on each data point $x_i$ of the minibatch.
The third step is the calculation of distances between all possible pairs within the minibatch.
We assume the output space of $f$ is equipped with a simple non-parametric distance function $d(\cdot,\cdot)$ such as Euclidean or cosine distance.
We denote by $\mathbf M$ the pairwise distance matrix of the minibatch: $M_{ij}=d(z_i, z_j)$. ~\footnote{
In some methods, not all elements of $M$ are taken, if we want to select some we can multiply $M$ by a mask.}

Finally, the last step is the computation of the loss from matrix $\mathbf M$, detailed in the next subsections, and which we minimize using bare SGD updates without momentum, for simplicity.

\subsection{Pairwise loss functions}
\label{subsec:loss_funcs}

Here we introduce the two pairwise loss functions which we use below in the definition of the batch level loss function.
At this point there is no notion of balance yet, it will be introduced in~\ref{sec:balance}.

\subsubsection{Contrastive margin loss}

For any batch indices $1 \leq i \neq j \leq b$, the contrastive margin loss~\cite{hadsell2006dimensionality} on the pair of descriptors $(z_i, z_j)$ has the form
\begin{equation}
	\ell_{i,j} = \mathds 1_{y_i=y_j} \ell_p(z_i, z_j) + \mathds 1_{y_i \neq y_j} \ell_e(z_i, z_j)
\end{equation}

It is either a \textit{positive} ($\ell_p$) or an \textit{entropy} ($\ell_e$) term, where:
\begin{align}
    \begin{split}
	&\ell_p(z_i, z_j) = d(z_i, z_j)^q; \\
	&\ell_e(z_i, z_j) = \max(0, m - d(z_i, z_j))^q,
	\end{split}
	\label{eq:contrastive}
	\end{align}
$m>0$ is the margin and usually $d$ is the euclidean distance and the exponent $q$ is $1$ (used in this paper) or $2$.

\subsubsection{InfoNCE loss}

The InfoNCE loss~\cite{oord2018representation}, for Noise Contrastive Estimation, is a common loss for SSL.
It consists of a simple modification over the multi-class N-pair loss~\cite{sohn2016improved} used in DML,
namely the use of a temperature parameter $\tau$,
and also appears under different names with or without scaling in other works~\cite{chen2020simple,xiao2017joint,wu2018unsupervised}.
Given batch indices $1 \leq i \neq j \leq b$ corresponding to a positive pair,
the loss for descriptors $(z_i, z_j)$ and family of contrastive terms $(z_k)_{k\in K}$ has the form
\begin{equation}
    \ell_{i,j} = \ell_p(z_i,z_j)+\ell_e(z_i,(z_k)_{k\in K})
    \end{equation}
where the \textit{positive} term $\ell_p$ and the \textit{entropy} term $\ell_e$ are:
\begin{align}
    \begin{split}
    &\ell_p(z_i,z_j) = d(z_i, z_j)/\tau;\\ 
    &\ell_e(z_i,(z_k)_{k\in K}) = \log\sum_{k \in K}\exp \left(-d(z_i, z_k)/\tau \right).
    \label{eq:infoNCE_onepair}
\end{split}
\end{align}
 A usual choice for $K$ is $K=\{j\}\cup\{k \in [\![1, b]\!]\ | \ y_k \neq y_i\}$, meaning that for the positive pair $(i,j)$, the entropy term thus includes the \textit{negative pairs} involving $i$, as well as the pair $(i,j)$ itself. We use it in the sequel, but other choices are possible.
 The temperature hyper-parameter $\tau$ controls the contrastivity. Usually $d$ is the cosine distance.

\section{Batch aggregation of pairwise losses}
\label{sec:balance}

In the last subsection we described how the contrastive margin and InfoNCE losses apply on a pair $(z_i, z_j)$. A minibatch of size $b$ contains $b^2{-}b$ pairs partitioned into the set $\mathcal P{=}\{(i,j) {\in} [\![1,b]\!]^2 | y_i{=}y_j,i{\neq j}\}$ of $b$ positive pairs and $\mathcal E{=}\{ (i,j) {\in} [\![1,b]\!]^2 | y_i {\neq} y_j\}$ of $b^2{-}2b$ negative pairs\footnote{These counts assumes two instances per class in the batch, typical of SSL (see Fig.~\ref{fig:pipeline}), but can be easily generalised. Also, some approaches mask a portion of the negative pairs, which we omit here for clarity.}.

How should the total loss on the minibatch be defined?
For the contrastive margin loss, the most obvious possibility is to take the \textit{global} average loss on all pairs as is usually done for element-wise losses.
The drawback is that the relative contribution of positive versus negative pairs to the total loss then depends on the batch size.
Another possibility is to average \textit{separately} for the negative and positive parts, in which case the relative contributions do not
depend on batch size~\footnote{although this option is the most natural from a normalization point of view, almost none of the common image retrieval
implementations that we are aware of use it, making them brittle to the choice of batch size, and encouraging back-door
means of modifying the balance for the `wrong' batch sizes, such as changing the number of instances per class. }, but may still be suboptimal.
For the InfoNCE loss, this question does not arise because the relative contributions of the positive versus entropy terms
are already defined in each $\ell_{i,j}$, but their ratio may still not be optimal.

In the following, we introduce the overall batch loss for both the constrastive margin loss and the InfoNCE loss as a combination of two sub-losses $\bar\ell_p(\mathbf z)$ and $\bar\ell_e(\mathbf z)$, and introduce a balance hyper-parameter (which may depend on the batch size):
\begin{align}
    \mathcal L(\mathbf z) &= \lambda_p \bar\ell_p(\mathbf z) + \lambda_e \bar\ell_e(\mathbf z),
    \label{eq:contrastive_decomposition}
    \\
    d\theta(\mathbf z) = \alpha \nabla_\theta \mathcal L(\mathbf z) &= \alpha \lambda_p \nabla_\theta \bar\ell_p(\mathbf z) +
    \alpha \lambda_e \nabla_\theta \bar\ell_e(\mathbf z).
    \label{eq:contrastive_decomposition_update}
    \end{align}
We see that the dynamics of the optimization problem can be uniquely characterized by two
hyperparameters: $\Lambda_p=\alpha \lambda_p$, $\Lambda_e=\alpha \lambda_e$, plus possibly the batch size $b$ if $\lambda_p,\lambda_e$ are taken to be dependent on it.

\subsection{Contrastive margin loss}
\label{ssec:ctrm_norm}

We define $\bar\ell_p$ and $\bar\ell_e$, \textit{positive} and \textit{entropy} terms, as the average values of (the contrastive margin loss versions of) $\ell_p$ and $\ell_e$ over the positive (resp. negative) pairs in the batch. With $\mathcal U$ denoting the uniform distribution, we have:
\begin{align}
    \begin{split}
    \bar\ell_p(\mathbf z) &= \mathbb{E}_{(i,j)\sim\mathcal{U}(\mathcal P)}[\ell_p(z_i, z_j)], \\
    \bar\ell_e(\mathbf z) &= \mathbb{E}_{(i,j)\sim\mathcal{U}(\mathcal E)}[\ell_e(z_i, z_j)].
    \label{eq:contrastive_means}
    \end{split}
\end{align}

Two settings are often used in practice, and both can be seen as special cases of our generic approach:
\begin{itemize}
    \item The \textit{global}-average minibatch loss assigns the same weight to positive and negative pairs, so that $\mathcal L {=} \frac{1}{|\mathcal P\cup\mathcal E|}(\sum_{\mathcal P} \ell_p {+} \sum_{\mathcal E} \ell_e)$. Since $|\mathcal E|{=}b^2{-}2b$ and  $|\mathcal P|{=}b$, this is the special case of Equation~\eqref{eq:contrastive_decomposition} where $\lambda_p=\frac{1}{b-1}$ and $\lambda_e=1-\frac{1}{b-1}$.
    \item Similarly, the \textit{separate}-average minibatch loss corresponds to the case $\lambda_p=\lambda_e=1$
\end{itemize}
In general, there is no reason to believe that any of these two hyperplanes in the space of hyper-parameters $\langle\Lambda_p,\Lambda_e,b\rangle$ gives optimal balance (\cf Figure~\ref{fig:hp_omniglot}).

\subsection{InfoNCE loss}
\label{subsec:infonce_loss}

Here, by analogy with Equation~\eqref{eq:contrastive_means}, we define $\bar\ell_p$ and $\bar\ell_e$ as average values of (the infoNCE loss versions of) $\ell_p$ and $\ell_e$, respectively, except the averages are taken over the positive terms only:
\begin{align}
    \begin{split}
    \bar\ell_p(\mathbf z) &= \mathbb E_{(i,j)\sim\mathcal{U}(\mathcal P)} [\ell_p(z_i,z_j)]; \\
    \bar\ell_e(\mathbf z) &= \mathbb E_{(i,j)\sim\mathcal{U}(\mathcal P)} [\ell_e(z_i,(z_k)_{k\in K_{ij}})],
    \label{eq:infoNCE_means}
    \end{split}
    \end{align}
where $K_{ij}=\{j\}\cup\{k\in[\![1,b]\!] \,|\, y_k\not=y_i\}$.

In practice, the usual InfoNCE loss is defined as $\mathcal L=\frac{1}{|\mathcal P|}\sum_{(i,j)\in\mathcal P}\ell_p(z_i,z_j)+\ell_e(z_i,(z_k)_{k\in K_{ij}})$ which is the special case of Equation~\eqref{eq:contrastive_decomposition} where $\lambda_e=\lambda_p=1$. As in the contrastive loss case, there is no reason to believe that this is the optimal configuration.

\section{A method for hyperparameter search}
\label{sec:hp_search}

Let $h{=}(\Lambda_p, \Lambda_e, b)$ be the hyper-parameters of our model, and $h {\rightarrow} \mathcal M(h)$ the evaluation metric which associates to a hyperparameter
configuration the value of the expected true evaluation performance, i.e. the test performance using the real evaluation measure, not its surrogate loss.

In the previous subsection we decomposed each contrastive loss into a two-task problem, and parameterized the optimization problem with hyperparameters $h=(\Lambda_p, \Lambda_e, b)$.
In the following we commonly refer to the value of $h$ as the `balance'.
Wang and Isola~\cite{wang2020understanding} highlighted the interest of tuning part of the balance (the relative contributions of
the positive to the entropy term) in the sub-case of SSL with InfoNCE .
In this section, we suggest that this tuning should also include the learning rate and batch size, and also applies to the contrastive margin loss.
Furthermore, we consider how to find the best setting of $h$ in the least number of training runs and suggest
a hyperparameter optimization (HPO) method for that purpose.

\subsection{Intuition}
\label{ssec:intuition}

We start by visualizing the form of $\mathcal M$ in a fixed-range cubic subspace $H$ of the hyper-parameter space.
For this, we use the \textit{Omniglot} dataset in an image-retrieval setting. 
\textit{Omniglot} is a handwritten character dataset often used for studying few-shot learning,
and which we find is also well-suited to studying image retrieval.
It contains 1623 different classes where each class has 20 examples.
Each example is a binary image (stroke data is also available but we do not use it in this work).
We train a ResNet with a constrastive margin loss, and plot R-mAP on the test set.

Figure~\ref{fig:hp_omniglot} shows the result grid for different values of $h$.
The first four batch sizes show an almost equal best performance of 0.85 R-mAP, although best performance then slightly degrades for larger batch sizes.
The grey (resp. beige) diagonal lines represent the 2D subspace accessible without an additional balance hyperparameter and using a \textit{global-average} loss (resp. \textit{separate-average} loss).
The grey diagonal lines change position with batch size because the relative proportion of positive versus entropy terms changes.
It seems that only one grey diagonal line and none of the beige diagonal lines include the best configuration for its batch size.

HPO-wise, two main qualitative observations can be made from this grid.
First, the performance landscape seems to be relatively simple, and approximately convex around the optimum.
Consequently, it seems sufficient to only exploit, and not explore, during the HPO.
In other words, searching only locally around the current best configuration will eventually lead to the best possible configuration.
Second, it seems that the 3D pattern has one main direction $d_0$, corresponding to varying $\Lambda_p$
and $\Lambda_e$ jointly, which is why it may be interesting to focus first on tuning \textit{orthogonal to this direction}.
Also, in the chosen hyper-parameter range, the value of the batch size does not seem so important.
Incidentally, the direction which is orthogonal to $d_0$ and keeps $b$ constant is exactly the \textit{balance} direction proposed
in~\cite{wang2020understanding}.

\subsection{HPO as reparameterized coordinate descent}
\label{subsec:reparam}

The intuition developed in the last subsection suggests a simple HPO algorithm in $H$: reparameterized coordinate descent.
We now detail this algorithm, also available in pseudo code in supp. mat. (Algorithm~\ref{algo:cd}).

Let $A$ be a $3\times 3$ linear reparameterization matrix the rows of which are the directions of a new coordinate system in $\log H$ in which we will perform coordinate descent.
We propose to simply use:

\begin{equation}
    A=
        \begin{pmatrix}
            -1 & 1 & 0\\
            1 & 1 & 0\\
            0 & 0 & 1
        \end{pmatrix},
    \label{eq:matrix_a}
    \end{equation}
which corresponds to the directions intuited just before: the first row is the \textit{balance} direction, the second row
corresponds to varying the \textit{joint learning rate}, and the third row is the \textit{batch size}. The reparameterisation $h\mapsto r$ is given by:
\begin{equation}
    \begin{gathered}
        r = A \log h^T
        \quad \text{with} \quad
        h = (\Lambda_p,\Lambda_e,b)
    \end{gathered}
    \label{eq:reparam}
    \end{equation}

This new coordinate system is also visualized in Appendix~\ref{sec:viz_a} (in supp. mat.). Working in it, coordinate descent consists of successive line searches alternating on the three directions.
Each line search consists of finding the optimal configuration on the current line.
An interesting property is that if metric $\mathcal M$ is close to its quadratic (second order) approximation (assumed positive, hence convex), and if the directions $A$ correspond to the eigenvectors of the Hessian
$\frac{\partial^2 \mathcal M}{\partial r^2}$, then coordinate descent can reach the optimum of $\mathcal M$  in only three full line searches. Hence the importance of the choice of $A$ in this ideal case.
Of course, in actual experiments $\mathcal M$ is not even perfectly convex so three line searches may not be enough, but it is still worth 
respecting the eigendirections of $\mathcal M$ near its optimum.

\section{Experiments and results}
\label{sec:exps}

In order to evaluate the importance of balancing contrastive losses, we perform experiments in DML and SSL.

\subsection{Datasets and metrics}

For DML, we use three datasets. Apart from \textit{Omniglot} already described, 
\textit{Stanford Online Products}~\cite{oh2016deep} is a dataset of approximately 120k images crawled from the web,
belonging to approximately 22k classes, making it larger-scale than Omniglot. 
For these two first datasets we report the mean average precision calculated over the subset of top-R candidates (R-mAP)\footnote{this is different from mAP over all candidates with a sum truncated to R, see supp. mat. }.
Finally, \textit{Landmarks-clean} is another image retrieval dataset cleaned by Gordo et al~\cite{gordo2016deep} on top of the original \textit{Landmarks}~\cite{babenko2014neural} dataset. 
\textit{Landmarks-clean} contains approximately 42k images from 586 classes.
Networks trained on \textit{Landmarks-clean} are usually
evaluated on the \textit{Roxford5k} or \textit{RParis6k} datasets~\cite{radenovic2018revisiting}. We report mean average precision (mAP) on the `medium' \textit{Roxford} subset.

For SSL, we train and evaluate our models on the \textit{Tiny-ImageNet} dataset, which is a smaller version of the famous ImageNet.
It contains approximately 110k labeled images from 200 classes.
Public test images do not have associated labels, which is why the common setup for this dataset is to use the public validation set as test set.
For evaluation, SSL networks are frozen, and a linear classifier is trained on top of them using training labels.
The classification accuracy of the resulting pipeline is then reported on the test set.
Because the dataset is quite small, the accuracy numbers on this dataset are low compared to SSL on full ImageNet, or compared to semi-supervised results
which use a few labeled examples.

In order to check that we start from a sound base, we checked our implementation against some reference numbers.
Table~\ref{table:baselines} (in supp. mat.) shows some performance numbers on three datasets
compared to reference implementations using the same loss.

\subsection{Implementation details}

\begin{figure}
    \centering
    \includegraphics{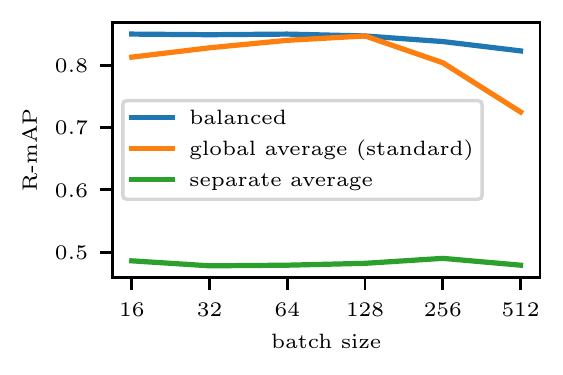}
    \caption{Maximum performance associated to using a balanced (blue) vs unbalanced (orange and green) loss, for a network trained with contrastive margin loss on Omniglot.
    Balancing the contrastive loss makes it possible to maintain near-maximal performance over a larger range of batch sizes.
    The \textit{global} and \textit{separate} averages in unbalanced losses refer to normalization options described in~\ref{ssec:ctrm_norm}. }
    \label{fig:balance_gain}
    \end{figure}

\begin{figure}
    \centering
    \includegraphics[width=0.35\textwidth]{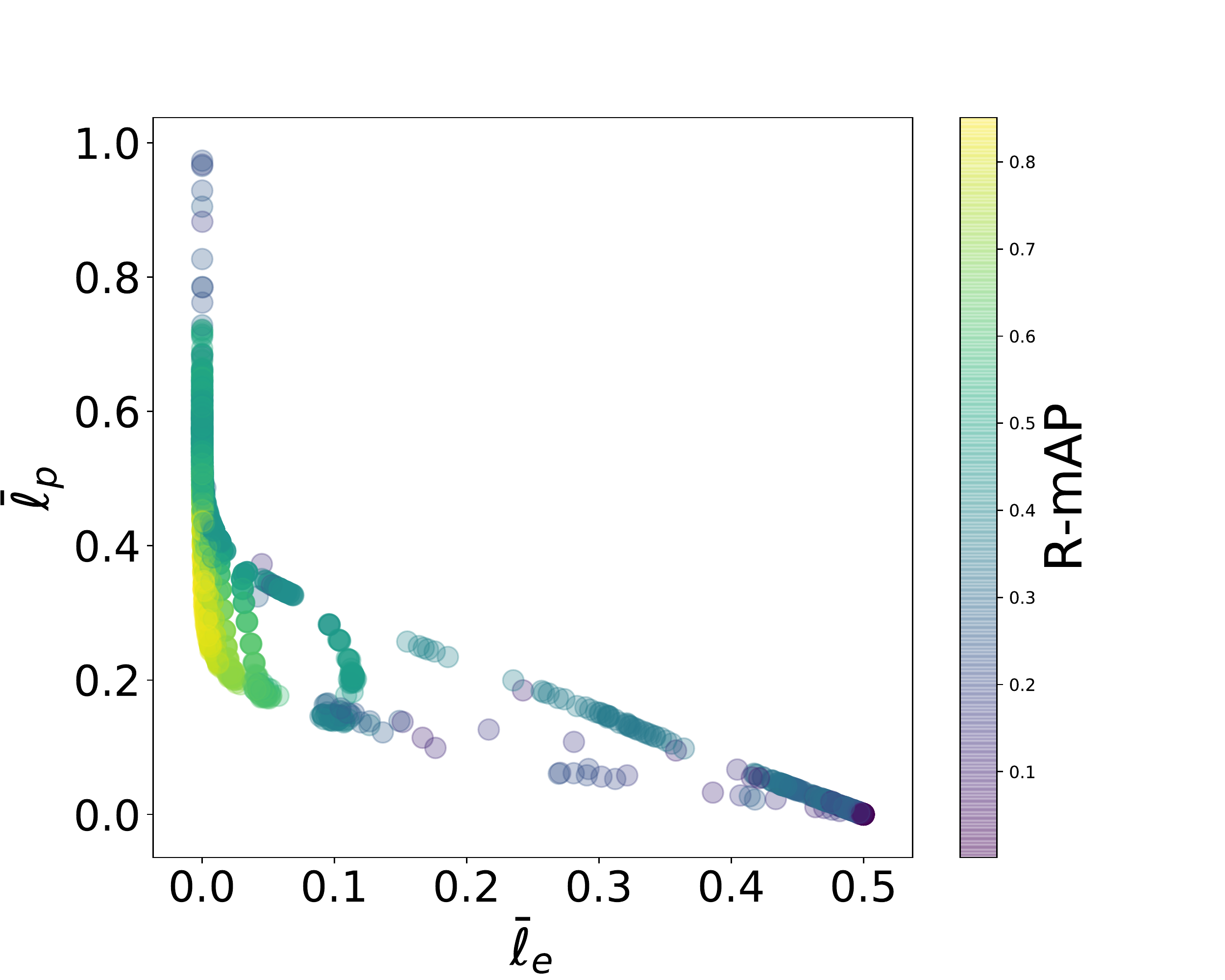}
    \caption{Scatterplot of R-mAP metric coordinated by positive and entropy losses after training, for Omniglot image retrieval (all batch sizes).
             Using a balanced loss makes it possible to best minimize \textit{both} the positive and the entropy subparts of the loss, which is required
             for good retrieval performance.
    }
    \label{fig:sublosses}
\end{figure}

\begin{figure}
    \centering
    \includegraphics{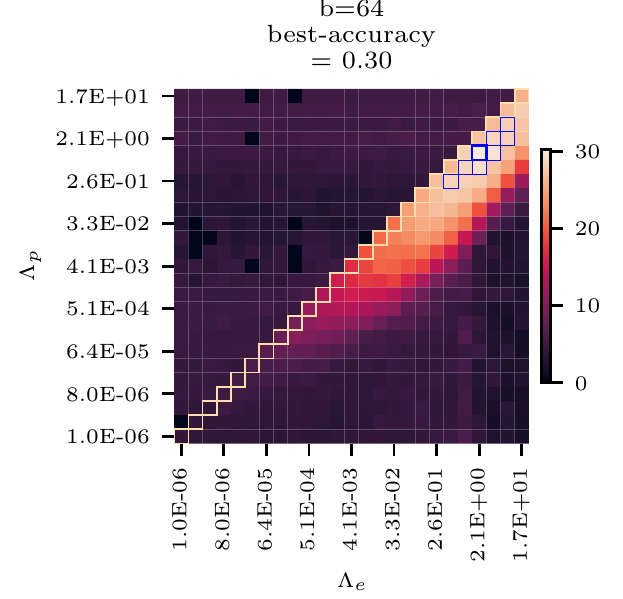}
    \caption{For a batch size $b=64$, color plot of TinyImageNet evaluation performance as a function of balance parameters.
    The test metric is linear classification performance on top of SSL features learned on TinyImageNet with an InfoNCE loss.
    Again we observe that \textbf{the best configuration is out of the main diagonal} reachable using a standard loss balance.
    The best accuracy with default balance is 0.28, with tuned balance it is 0.30. Tuning for the balance brings a 7.8\% relative improvement. }
    \label{fig:hp_tinyimagenet_infonce}
    \end{figure}

\textbf{Sampling.} Roth et al.~\cite{roth2020revisiting} showed that, for deep metric learning, a simple $2$-per-class strategy is sufficient.
Given batch size $b$, it consists in randomly selecting $b/2$ unique classes, from each of which $2$ samples are randomly chosen.
In SSL, using two transform views for each dataset image is also the standard approach.
For our experiments, we follow this $2$-per-class strategy.

\textbf{Networks and training.} For all DML experiments, we use a standard ResNet-18 with GeM pooling~\cite{radenovic2018fine}. SSL experiments are done with a ResNet-32 and the same data augmentations as SimCLR~\cite{chen2020simple}. 
For losses, we use $m=0.5$ for the margin and $\tau=0.1$ for the temperature.
All networks are trained with stochastic gradient descent, iterating for 3M steps for \textit{Omniglot},
2M steps for \textit{Stanford-Online-Products}, 1M steps for \textit{Landmarks-clean}, and 36M steps for \textit{Tiny-ImageNet}.
During training, the model is evaluated on a held-out validation subset of the training set.
The model version which peaks on the validation metric is selected and evaluated on the test set.

\textbf{HPO ranges.}
For HPO experiments, we vary $\Lambda_p$ and $\Lambda_e$ between $10^{-6}$ and $17$, by multiplicative increments of $2$.
For \textit{Omniglot} 3D experiments we also vary $b$ between $16$ and $512$ by multiplicative increments of $2$,
while for HPO experiments on other datasets we fix $b$ to 64 and only vary $\Lambda_p$ and $\Lambda_e$, effectively presenting results on a 2D grid.
Indeed, we cannot afford to build the whole 3D grid for all datasets (as an indication, reported experiments already make for approximately 3 years of GPU compute),
and the 3D-grid results on \textit{Omniglot} displayed on Figure~\ref{fig:hp_omniglot} show only a small influence of batch size on best attainable performance.

\textbf{HPO baselines.}
We compare the proposed HPO to three baselines: Random Search~\cite{bergstra2012random}, Tree-structured Parzen Estimators (TPE)~\cite{bergstra2011algorithms}, and Covariance Matrix Adaptation Evolution Strategy (CMA-ES)~\cite{loshchilov2016cma}.
Random search is a very simple yet solid baseline, while TPE and CmaEs are two classic model-based algorithms known for their performance.
TPE is a Bayesian algorithm which is typically recommended when the number of trials is low and the hyperparameters are not correlated.
CMA-ES models the covariance between hyperparameters, which is why it is recommended when a correlation between hyperparameters is expected,
and the number of trials is higher (such as $>100$, although for our 2D grids we find it to also outperforms TPE for small numbers of trials,
probably because hyperparameters are strongly correlated).
All three algorithms are available in Optuna~\cite{optuna_2019}.

\subsection{Interest of the balance hyperparameter}

\begin{figure*}[h]
    \centering
    \subfigure[]{\includegraphics[width=0.24\textwidth]{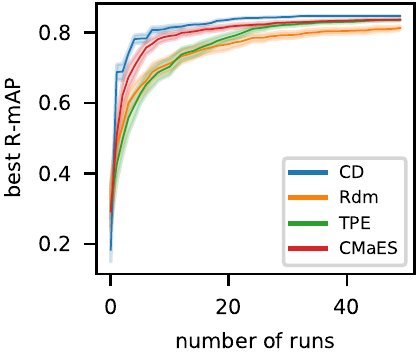}}
    \subfigure[]{\includegraphics[width=0.24\textwidth]{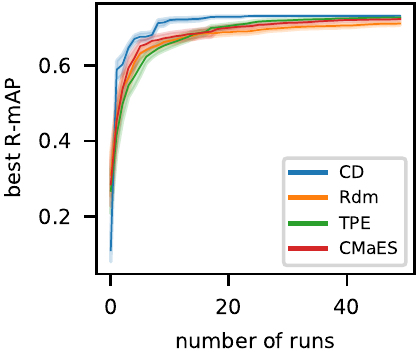}}
    \subfigure[]{\includegraphics[width=0.24\textwidth]{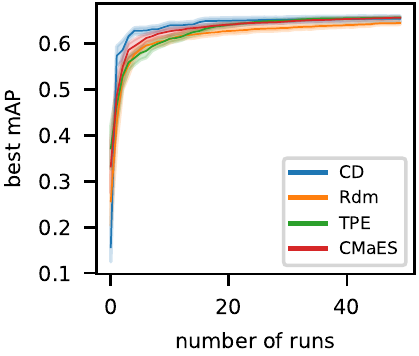}}
    \subfigure[]{\includegraphics[width=0.24\textwidth]{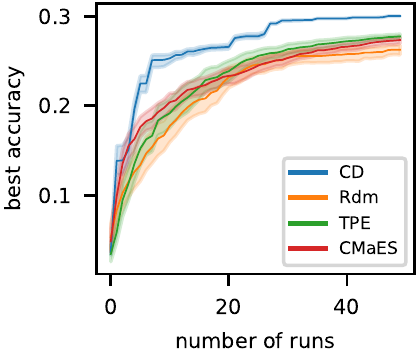}}
    \caption{HPO-2D curves towards obtaining the best possible balance for (a) Omniglot  (b) SOP  (c) Landmarks-Roxford5k  (d) TinyImagenet.
            These curves show the best performance reached as a function of the number of runs. Our coordinate descent method is designated by `CD'.
            Networks are trained using the loss which is most classic for the task: contrastive margin loss for DML tasks (Omniglot, SOP, Landmarks), and InfoNCE for SSL (TinyImageNet). }
    \label{fig:hpo_curves}
\end{figure*}

The first question we want to answer is what is the performance benefit of introducing an additional hyperparameter
for tuning the balance? Figure~\ref{fig:balance_gain} shows that by using an adequate loss balance, performance can be maintained over a larger range of batch sizes.
As already discussed in~\ref{ssec:intuition}, for a contrastive margin loss in general the best configuration is not reachable by tuning the learning rate only with fixed batch size.
Note that if a \textit{global-average} is used, tuning the batch size implicitely changes the balance, which is why
the problem is not so bad.
In this case, independently tuning the batch size and the learning rate is a very reasonable starting point.
Not doing so and tuning only the learning rate for fixed batch size results in a performance drop of 3.4\% R-mAP averaged on all batch sizes.
Figure~\ref{fig:sublosses} illustrates how balanced optimized losses $\bar{l}_p$ and $\bar{l}_e$ correlate with better retrieval results on Omniglot. 

With a standard unmodified InfoNCE loss, the problem is worse because the balance does not change as a function of batch size.
Figure~\ref{fig:hp_tinyimagenet_infonce} shows test accuracy results for a network trained with $b=64$ on TinyImageNet using an InfoNCE loss.
The beige diagonal corresponds to the subspace which is accessible without the additional hyperparameter.
This diagonal does not include the best configuration.
Using an additional hyperparameter and tuning increases the test accuracy from 0.28 to 0.30, or a 7.8\% relative improvement.
Unlike for the contrastive margin loss this improvement is expected to be consistent across batch sizes.

\subsection{HPO results}

When comparing HPO methods, we care both about the best performance a method reaches, and how fast it reaches it.
Figure~\ref{fig:hpo_curves} and Table~\ref{tab:hp_res} show such HPO-2D results using our coordinate-descent based
method compared to the baselines introduced before.
The different methods are compared in terms of value of the area under the curve (AUC)
of the best configuration reached as a function of the number of runs, and \textit{n-95}, representing the number of runs required to reach 95\% of the performance reached by the best HPO setting over all methods.
For each HPO method, reported numbers are obtained from the average of 80 HPO trajectories.
Performance values for continuous settings of $h$ are interpolated from a precomputed grid with sufficient resolution.
We observe consistent improvements over the three baselines across datasets and across the two losses.
Note that given the relatively low number of datasets we did not tune the initial relative budgets per direction and the line search
directions themselves.

\begin{table*}[]
    \small
    \setlength{\tabcolsep}{4pt}
    \centering
\begin{tabular}{@{}lcccccccccccc@{}}
\toprule
\multicolumn{1}{c}{\textbf{}} & \multicolumn{3}{c}{\textbf{Omniglot}}                                              & \multicolumn{3}{c}{\textbf{SOP}}                                                   & \multicolumn{3}{c}{\textbf{Landmarks-Roxford5k}}                                   & \multicolumn{3}{c}{\textbf{Tiny-ImageNet}}                                         \\ \midrule
                              & \multicolumn{1}{l}{AUC@10} & \multicolumn{1}{l}{AUC@20} & \multicolumn{1}{l}{n-95} & \multicolumn{1}{l}{AUC@10} & \multicolumn{1}{l}{AUC@20} & \multicolumn{1}{l}{n-95} & \multicolumn{1}{l}{AUC@10} & \multicolumn{1}{l}{AUC@20} & \multicolumn{1}{l}{n-95} & \multicolumn{1}{l}{AUC@10} & \multicolumn{1}{l}{AUC@20} & \multicolumn{1}{l}{n-95} \\ \midrule
Random                        & 0.58                       & 0.67                       & 41                       & 0.57                       & 0.63                       & 26                       & 0.54                       & 0.58                       & 18                       & 0.13                       & 0.17                       & $>$50                      \\
TPE                           & 0.62                       & 0.69                       & 24                       & 0.57                       & 0.63                       & 17                       & 0.53                       & 0.58                       & 14                       & 0.13                       & 0.17                       & $>$50                      \\
Cma-Es                        & 0.66                       & 0.73                       & 15                       & 0.58                       & 0.63                       & 18                       & 0.56                       & 0.6                        & 9                        & 0.17                       & 0.19                       & $>$50                      \\
\textbf{CD (ours)}            & \textbf{0.71}              & \textbf{0.77}              & \textbf{7}               & \textbf{0.61}              & \textbf{0.67}              & \textbf{8}               & \textbf{0.57}              & \textbf{0.61}              & \textbf{4}               & \textbf{0.19}              & \textbf{0.22}              & \textbf{27}              \\ \bottomrule
\\
\end{tabular}
        \caption{HPO results on different datasets. When comparing HPO methods, we care both about the best performance a method reaches, and how fast it reaches it.
        The different methods are compared in terms of value of the area under the curve (AUC)
of the best configuration reached as a function of the number of runs, and \textit{n-95}, representing the number of runs required to reach
    95\% of the performance reached by the best HPO setting over all methods. }
    \label{tab:hp_res}
\end{table*}

\begin{table}[]
    \small
\resizebox{\columnwidth}{!}{%
\begin{tabular}{@{}lccc@{}}
\toprule
\multicolumn{1}{c}{} & \textbf{Omniglot} & \textbf{SOP}  & \textbf{Landmarks-ROxford5k} \\ \midrule
$A_O$ (3, 3)   & \textbf{0.77}     & \textbf{0.67} & \textbf{0.61}                \\
$A_I$ (3, 3)            & 0.69              & 0.65          & 0.59                         \\
$A_O$ (7, 7)            & 0.58              & 0.58          & 0.58                         \\
$A_O$ (15, 15)          & 0.71              & 0.60           & 0.55                         \\
$A_O$ (3, 3) rev        & 0.72              & 0.64          & 0.58                         \\ \bottomrule
\\
\end{tabular}%
}
\caption{Ablation study showing the influence of directions in $A$, initial budget on each direction before switching direction (total bugdet is fixed to 50), and order of directions on AUC@20.
$A_O$ refers to the reparameterization matrix proposed in~\ref{subsec:reparam} while $A_I$ refers to using the identity matrix for $A$, in other words
no reparameterization. (i,i) numbers refer to the budget on each direction. `rev' refers to reversing which direction is searched first.
    See~\ref{ssec:intuition} for an intuition why the default setting $A_O$ (3, 3) works best.}
    \label{tab:ablation}
    \vspace{-3mm}

\end{table}

\subsection{Ablation studies}

Finally, we did a basic study of the influence of reparameterization directions and budgets.
Table~\ref{tab:ablation} shows a summary of associated results.
Compared to using original balance directions corresponding to using the identity matrix for $A$, the proposed directions offer an improvement.
This confirms the interest of working in eigendirections of the HPO space when using coordinate descent.
Regarding initial budget per direction, we observe that a relatively low bugdet such as 3 trials per direction is the most effective choice for the
beginning of the HPO.
Also, it is more effective to start with the balance direction first.
This minimal tuning also shows that \textit{the same setting} for $A$, budget and order work best \textit{for all datasets}.
This setting should be viewed as a default hyper-hyperparameter which does not need to be tuned,
similar to parameters of TPE and CmAES which are generally not tuned.

\section{Discussion and perspectives}

The results presented in this paper show that it is beneficial to balance contrastive losses.
They also raise a number of points which we briefly explain here.

First, we perform experiments with minibatches organized by pairs of labels ($k$-per-class sampling with $k=2$).
The number of positive interactions in $\mathbf M$ is $b(k-1)$.
Our discussion easily extends to $k$ values other than two, and we want to mention that multiple implementations of the contrastive-margin loss implicitely use different values of $k$ as an implicit way of tuning the balance.

The HPO method we introduced was evaluated and compared, assuming no knowledge of a reasonable starting configuration.
In practice, such a starting configuration is almost always available to scientists, and plugging it into our coordinate-descent HPO method
is particularly easy.
This will noticeably accelerate the search, unlike other HPO methods which are not obvious to warm-start.

We demonstrated that the directions in $A$ and relative tuning budgets affect HPO effectiveness, and we provided
experimentally reasonable values for them, but did not tune more on this limited number of datasets.
It may be possible to find slightly better and theoretically justified values for $A$ by studying the relationship between
batch size, learning rate, and generalization for each sub-loss.
Appendix~\ref{sec:neg} (in supp. mat.) presents an attempt towards this.
It would also be possible to approximate the Hessian of $\mathcal M$ in the optimal configuration through finer-resolution experiments.

Also, although it already beats other methods, our implementation of coordinate descent can still be improved.
Indeed, the bounded 1D golden section search performed in each line search is initialized with a search bracket corresponding
to the limits of the overall search space.
This prevents the overall HPO process from focussing first on a local neighbourhood
of the current best configuration.
Further performance improvement could probably result from slightly modifying
Algorithm~\ref{algo:line_search} (in the appendix) such that the bracket for each line search is initialized to a local neighborhood of the current best
configuration.

For a fixed loss (type and hyper-parameters), our results also seem to suggest that the best value of $h$ may not depend too much on the dataset (within a same task: see~\ref{ssec:best_hp_values}).
Of course, loss hyper-parameters such as margin or temperature will affect the balance, but for fixed values of these
it seems that practitioners could simply reuse pre-tuned values of $h$.

Another confounding factor is batch normalization.
It is strongly suspected that batch normalization behaves as an entropy term~\cite{bn2020}.
In this sense, BN hyper-parameters are expected to interfere with the entropy term.

Finally, our findings raise questions on the usefulness of common training practices in DML and SSL such as negative
mining and memory queues.
Such practices modify loss balance, and our work showed that balance crucially affects performance.
Therefore, it is reasonable to question to what extent negative mining and memory queues are useful \textit{per se}, as opposed to
\textit{indirectly} because they modify loss balance.
Appendix~\ref{sec:xbm} (in supp. mat.) presents an initial study 
on cross-batch memory~\cite{wang2020cross} where we observe that tuning
the balance \textit{reduces by a factor of two} the discrepancy between without/with cross-batch memory.
Further experiments will be required to substantiate this understanding.

\section{Conclusion}
\label{sec:conclusion}

In this study we analyzed the effect of loss balance on test performance for deep metric learning and self-supervised learning
using contrastive losses.
We decomposed each of two common contrastive losses into a positive term and an entropy term.
We demonstrated that loss balance has a strong effect on evaluation performance, and we provided a simple and effective HPO method
based on coordinate descent in a transformed hyperparameter space to tune it.

A number of directions remain open for future work.
The most direct one is to evaluate whether the interest of negative mining and memory queues holds when balance is
properly tuned.
The other main one is to build on these findings to develop a better understanding of what the exact best directions of the reparameterization matrix should be,
relating for each term, batch size, learning rate and generalization.
This could make possible a simple scaling rule for contrastive losses, in the same spirit as the ubiquitous \textit{linear scaling}~\cite{bottou2018optimization,you2017scaling,goyal2017accurate}
used between learning rate and batch size in the case of classification.


{\small
\bibliographystyle{ieee_fullname}
\bibliography{paper}
}

\clearpage

\appendix

\section{Performance numbers}

In order to situate our implementation compared to some reference ones, we briefly present some performance numbers
on three datasets in Table~\ref{table:baselines}.
Similar or slightly better numbers are obtained, although our implementation generally uses a smaller network
in order to save computation time in the HPO study.

\begin{table*}[]
    \small
    \centering
\begin{tabular}{@{}cccc|ccc|ccc@{}}
\toprule
\multicolumn{4}{c|}{\textbf{SOP} (CL)}        & \multicolumn{3}{c|}{\textbf{Landmarks-ROxford5k} (CL)} & \multicolumn{3}{c}{\textbf{Tiny-ImageNet} (InfoNCE)} \\ \midrule
                                  & Loss   & P@1  & R-mAP &                                      & Loss & mAP        &                                     & Loss    & acc  \\
\cite{oh2016deep}                 & CL     & 0.62 &   -   & \cite{revaud2019learning}-GeM-CL     & CL   & 0.65       &                                     &         &      \\
\cite{musgrave2020metric}         & CL     & 0.73 & 0.44  & \cite{radenovic2018revisiting}-R-GeM & CL   & 0.65       & \cite{patacchiola2020self}(SimCLR)  & InfoNCE & 0.26 \\
\cite{venkataramanan2021takes}-CL & CL     & 0.75 &   -   &                                      &      &            &                                     &         &      \\ 
\textbf{ours}                     & CL     & 0.73 & 0.44  & ours                                 & CL   & 0.66       & \textbf{ours}                       & InfoNCE & 0.30 \\ \bottomrule 
    \\
\end{tabular}%
\caption{Base performance numbers of our implementation compared to some reference numbers reported
    \textit{for the same loss} in other papers. 
    Note that architectures, optimizers, and pooling layer differ slightly between the different implementations.
    For SOP and Landmarks, we use ResNet-18, GeM pooling, and SGD,
    \cite{radenovic2018revisiting}-R-GeM and~\cite{revaud2019learning} use ResNet-101, GeM pooling, and Adam,
    \cite{musgrave2020metric} uses BN-Inception, a linear projection, and RMSProp,
    ~\cite{oh2016deep} use Google LeNet, a linear projection, and SGD,
    and~\cite{venkataramanan2021takes} use a Resnet50, a combination of average and max pooling followed by a linear
    projection, and AdamW.
    For TinyImageNet, we use ResNet-32 and SGD, and~\cite{patacchiola2020self} uses ResNet-32 and Adam.
}
    \label{table:baselines}
\end{table*}

\section{Coordinate descent algorithms}

Algorithm~\ref{algo:cd} details the coordinate-descent algorithm for HPO.
Algorithm~\ref{algo:line_search} details how each line search of the coordinate-descent is done.

\begin{algorithm}[h]
    \small
\SetAlgoLined
\SetKwInOut{Input}{input}\SetKwInOut{Output}{output}
\Input{starting point $h_0$ \\
    matrix of directions $A$ \\
    budgets per direction $(c_0, c_1, c_2)$ \\
    total budget $c$ \\
    search space $H=$\\ $(\Lambda_{p_{min}}, \Lambda_{p_{max}}, \Lambda_{e_{min}}, \Lambda_{e_{max}}, b_{min}, b_{max})$}
\Output{best configuration $h=(\Lambda_p, \Lambda_e, b)$}
Initialization: starting point and current direction \\
$h \leftarrow h_0$ \\
$i \leftarrow 0$ \\
 \While{not finished(c)}{
  Perform a line search of $c_i$ trials:
  $h \leftarrow \textit{line\_search}(h, A[i, :], c_i, H)$ \\
 $i \leftarrow (i+1) \% 3$ \\
 Optionally update budget:
 $c_i \leftarrow c_i(\textit{history})$
 }
 \caption{Coordinate-descent HPO}
    \label{algo:cd}
\end{algorithm}

\begin{algorithm}[h]
    \small
\SetAlgoLined
\SetKwInOut{Input}{input}\SetKwInOut{Output}{output}
\Input{starting point $h_l=(\Lambda_{p_l} \Lambda_{e_l}, b_l)$ \\
    search direction $a_l$ \\
    budget (number of trials) $c_l$ \\
    search space $H$}
\Output{best configuration found on this line: $h=(\Lambda_p, \Lambda_e, b)$}
Initialization: starting point $h \leftarrow h_l$, $\gamma=0$ \\
Equation of current line is $(\mathcal L): h_l + \gamma a$ \\
    Determine \textit{search\_bracket}$[\gamma_{min}, \gamma_{max}]=(\mathcal L) \cap H$ \\
 \For{$c$ runs}{
  Perform a golden-section step: \\
     $(\gamma_{min}, \gamma, \gamma_{max}) \leftarrow \textit{step}(\gamma_{min}, \gamma, \gamma_{max})$\\
    $h \leftarrow h_l + \gamma a$
 }
 \caption{(bounded golden-section) \textit{line$\_$search}}
    \label{algo:line_search}
\end{algorithm}

Because using a fixed budget per direction is inefficient as coordinate-descent converges to the best hyperparameter
value and the current line needs to be explored better, we use a simple slope-based rule to increase the budget per direction as needed:
when the slope of performance metric per step goes below a certain threshold (here 0.02) we simply multiply (by 2.) the budget per direction.

\section{R-mAP vs mAP@R}

Given a query image $\mathbf x_q$ and a list $\mathcal D_q = (\mathbf x_1, .., \mathbf x_M)$ of candidate images \textit{ranked by decreasing relevance},
the average precision for this query is defined by a sum over successive ranks $k$:

\begin{equation}
    AP(\mathbf x_q, \mathcal D_q) = \sum_{k=1}^M P(k) \Delta r(k),
    \end{equation}
where $P(k)$ is the precision at rank k in $\mathcal D_q$ and $\Delta r(k)$ is the difference in recall between ranks $k-1$ and $k$.
If we denote by $R_q(\mathcal D_q)$ the number of relevant candidate images for this query image and $\text{rel}(k)=\text{rel}(\mathbf x_k)=\mathds 1_{y_q=y_k}$ the relevance of
the $k$-th image, average precision can also be written as:

\begin{equation}
AP(\mathbf x_q, \mathcal D_q) = \frac{1}{R_q(\mathcal D_q)} \sum_{k=1}^M P(k) \text{rel}(k).
\end{equation}

In this paper, for the SOP and Omniglot datasets, where evaluation candidate images are shared between all query images,
for each query we simply calculate average precision over the subset of top-$R_q$ ranked images from $\mathcal D$:

\begin{align}
AP\text{-topR}&(\mathbf x_q,  \mathcal D) = AP(\mathbf x_q, D[1:R_q(\mathcal D)])) \\
&= \frac{1}{R_q(\mathcal D[1:R_q(\mathcal D)])} \sum_{k=1}^{R_q(\mathcal D)} P(k)\text{rel}(k).
\end{align}
The R-mAP metric is then the average of $AP\text{-topR}(\mathbf x_q, \mathcal D)$ over all queries.
Note that this metric is slightly different from the mAP@R metric introduced by~\cite{musgrave2020metric}, where AP@R is the
average precision on the \textit{full} dataset, but with a sum truncated to $R_q(\mathcal D)$, which results in a different denominator:

\begin{equation}
AP@R(\mathbf x_q, \mathcal D) = \frac{1}{R_q(\mathcal D)} \sum_{k=1}^{R_q(\mathcal D)} P(k)\text{rel}(k)
\end{equation}

\section{Additional experimental results}

\subsection{Switching the task and the loss}
\label{ssec:switching}

\begin{figure}
    \centering
    \includegraphics{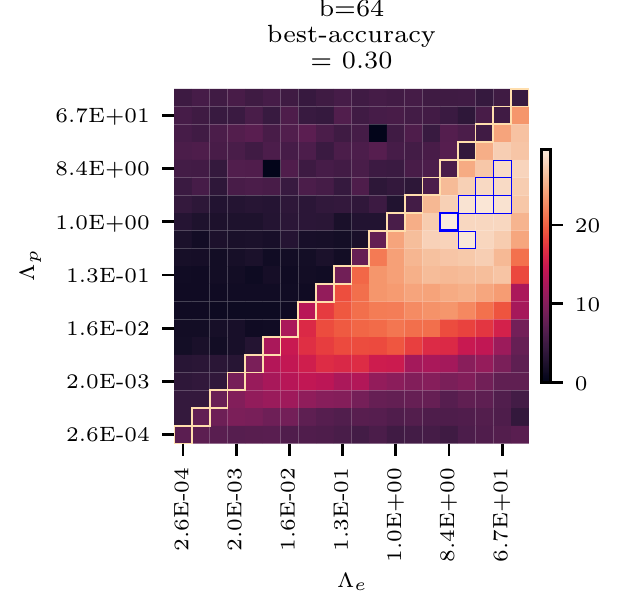}
    \caption{Classification accuracy of a network trained with SSL on Tiny-ImageNet with contrastive margin loss}
    \label{fig:ssl_ctr}
    \end{figure}

In our main experiments, we used the contrastive margin loss for DML tasks and the infoNCE loss for SSL, as is most classically done.
However, both losses can be used interchangeably on both tasks.
As an example, Figure~\ref{fig:ssl_ctr} shows the result of SSL training on Tiny-Imagenet using a contrastive margin loss.

\subsection{Best values of balance hyperparameters}
\label{ssec:best_hp_values}

It is interesting to ask whether the best values of the balance hyperparameters depend on the task and on the loss.
We observe that they mostly depend on the task: the best values for SSL experiments (pretraining) are relatively different from
the best values for DML experiments (fine-tuning of a pretrained model).
Within a same task, the best values are very close, and in our experiments we observed that there is very little
difference between the few-best settings.

\begin{table}[]
\resizebox{\columnwidth}{!}{%
\begin{tabular}{@{}llccc@{}}
\toprule
dataset                & task & loss    & best $\Lambda_p$ & best $\Lambda_e$ \\ \midrule
\textbf{Omniglot}      & DML  & margin  & 8e-3           & 2.             \\
\textbf{SOP}           & DML  & margin  & 1e-2           & 4.             \\
\textbf{Landmarks}     & DML  & margin  & 4e-3           & 5e-1           \\
\textbf{Tiny-ImageNet} & SSL  & margin  & 1.             & 8.             \\
\textbf{Tiny-ImageNet} & SSL  & infoNCE & 1.             & 2.             \\ \bottomrule
    \\
\end{tabular}%
}
\caption{Table showing the best balance parameters found for the different datasets for $b=64$.
        On this small set of experiments, it seems that the best configuration depends on the task
    (DML from a pretrained model vs SSL from scratch), but for a same task and same loss the balance can be re-used.
    In our experiment on SSL the best balance also did not depend on the loss (margin vs InfoNCE) but this will probably not
    hold with different loss hyperparameters. }
\end{table}

\section{Cross-batch memory}
\label{sec:xbm}

Cross-batch memory (XBM)~\cite{wang2020cross} is a technique which consists of augmenting the current minibatch with descriptors and labels (detached)
from previous minibatches. It is also known under different other names such as memory bank or memory queue~\cite{wu2018unsupervised,wang2020cross,xiao2017joint,he2020momentum,tian2019contrastive,misra2020self}.
In the implementation from \cite{wang2020cross} where the memory-augmentation is done one one `side' of the similarity matrix $\mathbf M$, using a memory of $K$ minibatches
approximately increases the number of negatives by $K$ and keeps the number of positives constant (if $bK << N/l$ with $l$ the number of instances per class).
This effectively amounts to using $\eta=K$ in equations from \ref{sec:balance}.
Therefore, the use of cross-batch memory changes loss balance.
As a consequence, it is reasonable to question whether the benefit of XBM comes from the mechanisms associated with the use of a memory (increased effective batch size,
inclusion of descriptors which were obtained with earlier versions of the model), or simply from modifying the balance of the contrastive loss.

In order to answer this question, we consider the three following setups on SOP with a contrastive margin loss: in (S1) we use XBM exactly as implemented by the authors: the XBM loss is progressively introduced
as memory is filled. Its negative part is a \textit{sum}, therefore the balance progressively changes as memory size increases, until memory is filled to its maximum length.
In (S2) and (S3), the balance of both the minibatch loss and the XBM loss is fixed to the same value.
For (S2), this value is the default value used by~\cite{wang2020cross} on the minibatch loss.
For (S3), this balance value is re-tuned using our coordinate-descent algorithm, separately for without/with XBM.
Note that for a fair comparison between (S1) and (S3), we also tried to tune the XBM line of (S1), but found that the balance is already optimal.
In all cases, if XBM is used is is introduced after 1000 steps, and the total loss is the standard minibatch loss plus the XBM loss. Training is done for 70k steps instead of 35k to ensure we do not miss the optimal point.

Table~\ref{table:xbm} presents the results obtained. First, we find that the basic performance of the no-XBM contrastive loss (S1/S2 without XBM) is slightly underreported (0.68 R@1 instead of 0.64).
Second, tuning the balance of the basic contrastive loss increases R@1 from 0.68 to 0.73. Finally, introducing XBM (either in its standard, balance-alterating version or in the fixed-tuned-balance version)
further pushes R@1 to 0.77.
This means that \textit{balancing the contrastive margin loss makes up for more than half of the reported discrepancy between with/without XBM}.

Interestingly, although our fixed-balance version of XBM is able to reach almost the same R@1 as the variable-balance original implementation, training to optimality
requires 1.5-2x more iterations.
This further questions the effect of variable-balance in XBM .

\begin{table}[]
\resizebox{\columnwidth}{!}{
\begin{tabular}{@{}lcc@{}}
\toprule
Balance                          & without XBM                  & with XBM             \\ \midrule
Original (XBM authors)           & 0.64 / -                     & \textbf{0.77 / -}    \\
(S1) Original (reproduced)            & \multirow{2}{*}{0.68 / 0.69} & \textbf{0.77 / 0.80} \\
(S2) Fixed to initial default balance &                              & 0.71 / 0.72          \\
(S3) Fixed and tuned                  & 0.73 / 0.75                  & \textbf{0.77 / 0.79} \\ \bottomrule
    \\
\end{tabular}%
}
\caption{Recall@1 results for training a ResNet-50 for Image Retrieval on Stanford Online Products with and without XBM, and
    for different settings of the loss balance. The two performance numbers refer to without/with learning rate decay.
    \textbf{Tuning the balance reduces by more than two times the discrepancy between with and without XBM. }
    Also, in the way XBM is originally implemented, balance implicitely changes during the training as XBM is introduced, resulting in a \textit{variable} balance.
    We are able to obtain almost the same performance with a tuned \textit{fixed} balance, although the training requires more training steps. }
\label{table:xbm}
\end{table}

\section{2D vs 3D HPO}

\begin{figure}
    \centering
    \includegraphics{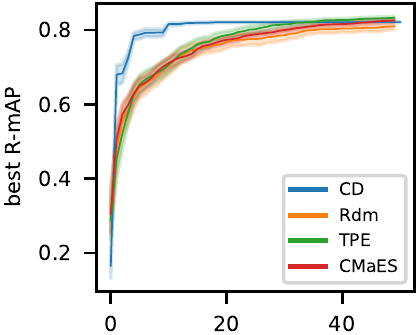}
    \caption{3D-HPO results on Omniglot}
    \label{fig:omniglot3D}
    \end{figure}

In the main part of the paper we present HPO results on 2D balance in the space $(\Lambda_e, \Lambda_p)$. As written in Equation~\ref{eq:reparam}, batch size $b$ can also be
included in this optimization. We chose not to do so for all datasets because on Omniglot (Figure~\ref{fig:hp_omniglot} we observed that the best configuration of $(\Lambda_p, \Lambda_e)$
depends very little on $b$, and building a full 3D-grid or doing multiple runs of 3D HPO on all datasets is required too much computational resources.
Figure~\ref{fig:omniglot3D} shows HPO curves on the Omniglot dataset (on which we have a 3D grid) for the different methods. In this case the budget for CD is set to (3, 3, 3).
In this case where one hyperparameter ($b$) is far less important than the other two, CD is even better than other methods.

The question of whether/how the best configuration of $(\Lambda_p, \Lambda_e)$ depends on $b$ in the general case for pair-based contrastive losses and how to correctly scale contrastive losses is still unclear, and to be further investigated,
we expect that the positive part should follow a linear relation $b_p \sim \Lambda_p$ as classically for pointwise losses~\cite{bottou2018optimization,devarakonda2017adabatch,smith2017bayesian,jastrzkebski2017three}).

\section{Visualization of matrix $A$}
\label{sec:viz_a}

\begin{figure}[h]
    \centering
    \includegraphics[width=\columnwidth,trim={3 3 3 2cm},clip]{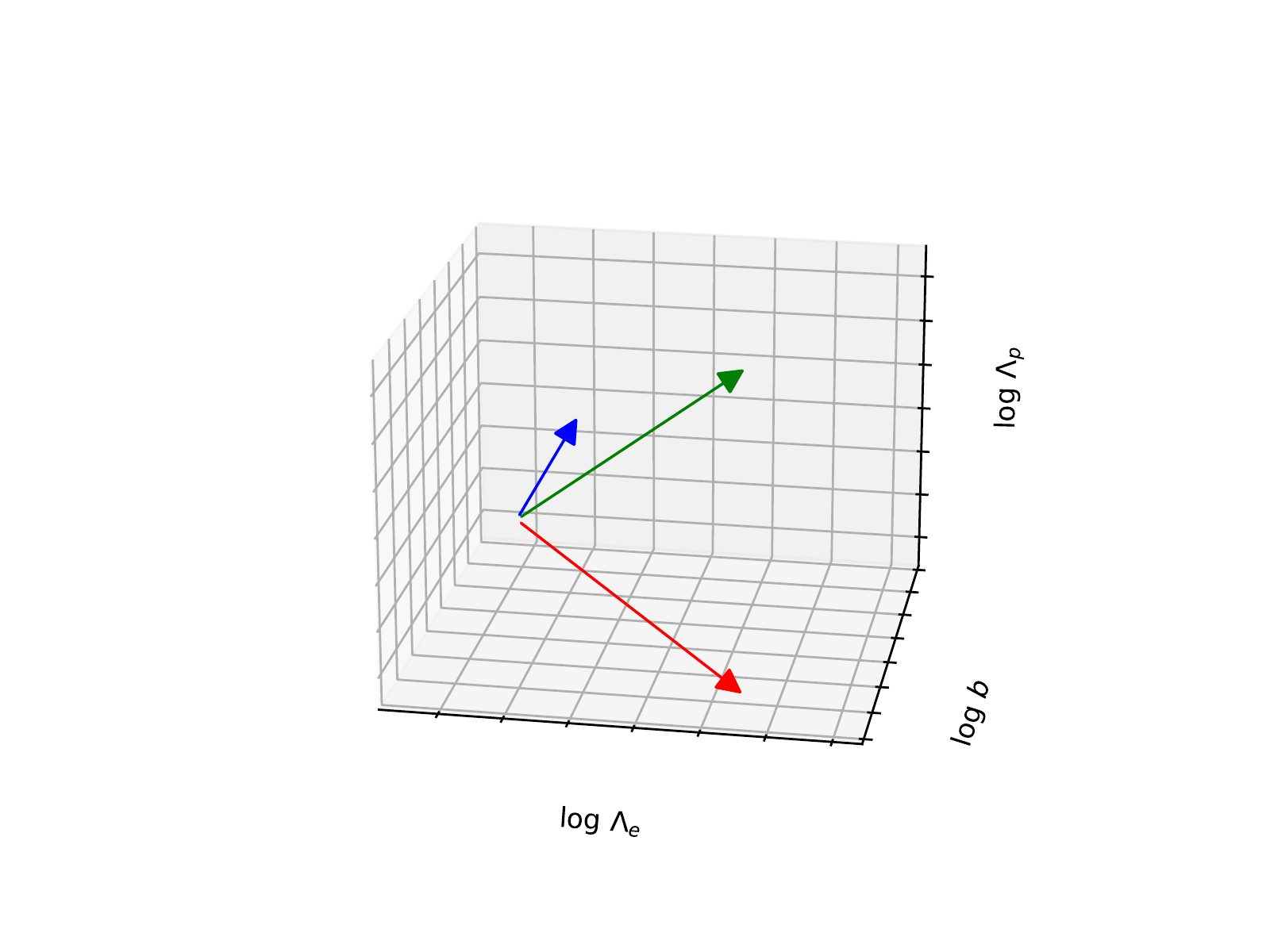}
    \caption{Visualization of the reparameterization directions proposed in~\ref{subsec:reparam}.
    We plot in the space of $\log h=(\log \Lambda_p, \log \Lambda_e, \log \b)$ the coordinate vectors
             of the new coordinate system, corresponding to the row of $A$: $\color{red}{A[0,:]}$, $\color{green}{A[1,:]}$, and $\color{blue}{A[2,:]}$}
    \label{fig:viz_matrix_a}
    \end{figure}

In order to better visualize the values the reparameterization matrix $A$ introduced in~\ref{subsec:reparam}, we plot here the
directing vectors of the new coordinate system in the space of $\log h$. The plot is displayed on Figure~\ref{fig:viz_matrix_a}.
Coordinates in this new system are denoted by $r$, where $r = A \log h^T$.

\section{Negative result: an attempt at theoretically deriving the best reparameterization directions}
\label{sec:neg}

We also questioned whether we could theoretically justify how the different balance hyperparameters should go together.
Drawing inspiration from the known scaling rule of batch size and learning rate for classification losses, we attempted to generalize this
to interaction losses, and deduce the reparameterization directions from this.
Although we did not succeed, we still present a negative result here, which we believe can be useful.

\subsection{Background: linear scaling of learning rate and batch size for non-interaction losses}
\label{ssec:bs_lr_background}

In stochastic optimization, learning rate and batch size are two parameters which should be chosen jointly rather than independently.
Consider a simple classification loss $\mathcal L(\mathbf z) = \frac{1}{b} \sum_{i=1}^b \ell(z_i)$ on a minibatch $\mathbf z=(z_0, .., z_b)$.
We denote $(\alpha_0, b_0)$ a setting of the learning rate $\alpha$ and number of samples $b$ which leads to the best evaluation performance.
Different works showed for SGD that within a range of learning rates $0<\alpha<\alpha_{\text{max}}$,any configuration $(\alpha,b)=(f \alpha_0, f b_0)$ leads to the same validation performance
and to the same optimization dynamics of $\mathcal L$ as a function of the \textit{number of data points seen},
because the update step at each iteration is multiplied by $f$, but the number of examples (and computation cost) per iteration
is also multiplied by $f$.
This was proved globally~\cite{bottou2018optimization} in the case of a strongly-convex loss with Lipschitz-continuous gradients,
and locally~\cite{devarakonda2017adabatch,smith2017bayesian,jastrzkebski2017three}.
The proof uses the fact that $\mathbb E(\nabla_\theta \mathcal L)$ does not depend
on the number of samples $b$, and that $\mathbb V(\nabla_\theta \mathcal L) \propto \frac{1}{b}$.
If the ratio $\alpha/b$ goes under $\alpha_0/b_0$, then overfitting will occur,
whereas if it goes over $\alpha_0/b_0$, then too much variance in update steps will prevent reaching the optimum with enough precision.
This scaling rule is also behind all large-batch training techniques~\cite{you2017scaling,goyal2017accurate}.

\subsection{Application to interaction losses}
\label{sec:constraints}

The main loss is a weighted sum over $\bar\ell_p$ and $\bar\ell_e$:
\begin{align}
\mathcal L(\mathbf z) &= \lambda_p \bar\ell_p(\mathbf z) + \lambda_e \bar\ell_e(\mathbf z)
\end{align}

If we view the DML or SSL loss as a two-task problem we can hypothesize that each of the sub-task separately should obey the scaling rule introduced in~\ref{ssec:bs_lr_background}.
In the following we deduce what follows if this is the case. We do this for the contrastive margin loss but the same reasoning can apply to InfoNCE (only the expression of the variance of the entropy term gradients will be different).

\subsubsection{Positive term}

For the positive loss to generalize as well as possible, the value of learning rate
and number of samples in $\bar\ell_p$ on a minibatch (which is $b(n-1)=b$ for the $2$-per-class strategy) should have the right ratio.
Denoting $r_p$ the value of this ratio, and noting that using a $\lambda_p$ multiplier on the loss is equivalent
to multiplying the SGD learning rate $\alpha$ by the same $\lambda_p$, we get

\begin{equation}
    (\mathcal C_p) \quad \frac{\Lambda_p}{b} = r_p
    \label{eq:cp}
    \end{equation}

\subsubsection{Entropy term}

The same constraint on learning rate vs number of samples applies on the entropy term, but this time the number of samples
is $b(b-n)\eta \sim b^2 \eta$.
Denoting $r_e$ the ideal ratio between equivalent learning rate and number of samples, we get the following constraint

\begin{equation}
    (\mathcal C_e) \quad \frac{\Lambda_e}{b^2 \eta} = r_e
    \label{eq:c3}
    \end{equation}

\subsubsection{Balance}

Finally, even if each sub-task is correctly parameterized in terms of batch-size versus learning rate ratio and generalization,
their relative importance in the total loss should be balanced in order to avoid overfitting the easier task whilst the more
difficult task is still learning~\cite{kendall2018multi,liang2020simple}.
Another way of viewing this constraint is the following: as long as both $\mathcal C_p$ and $\mathcal C_e$ are
satisfied, the ratio $\Lambda_e/\Lambda_p$ quantifies the relative advancement of each subproblem $\bar\ell_p$ and
$\bar\ell_e$ as a function of the \textit{number of iterations}).
We want to avoid the case where one-subproblem reaching its optimum whilst the other is still underfit.
Denoting $r_{eq}$ the ideal value of the ideal ratio, the constraint translates as

\begin{equation}
    (\mathcal C_{eq}) \quad \frac{\Lambda_p}{\Lambda_e} = r_{eq}
    \label{eq:ceq}
    \end{equation}

\subsubsection{Putting constraints together}

\begin{figure}
    \centering
    \includegraphics[width=0.8\columnwidth,trim={0 0 0 3cm},clip]{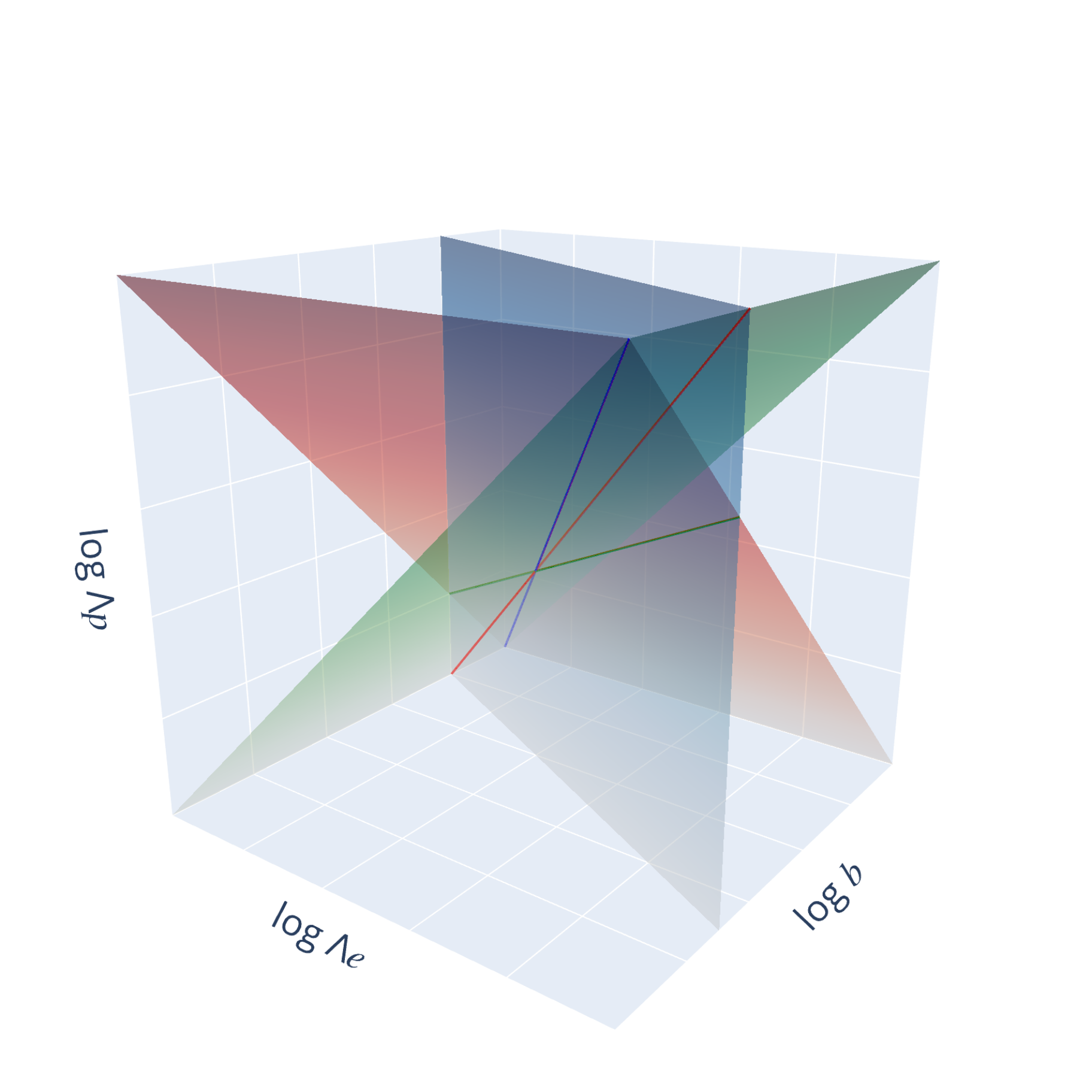}
    \caption{A 3D view of the three constraints on $h=(\Lambda_p, \Lambda_e, b)$.
        $\mathcal C_p$ is in red, $\mathcal C_e$ is in blue, and $\mathcal C_{eq}$ is in green. Each of these planes represent the set of points
        where the associated constraint is zero. In order to decouple the search (coordinate descent), it is natural to perform coordinate descent
        along directions which vary only one constraint. }
    \label{fig:constraints}
    \end{figure}

Taking the log on each constraint equation, we get a linear system $A_{th} \log h^T = \log r_{c}$ of three contraints in log-configuration
space, with

\begin{equation}
    \begin{gathered}
        A_{th}=
        \begin{pmatrix}
            1 & 0 & -1\\
            0 & 1 & -2\\
            1 & -1 & 0
        \end{pmatrix}
        \ , \\
        \log h^T=
        \begin{pmatrix}
            \log\Lambda_p\\ \log\Lambda_e\\ \log b\\
            \end{pmatrix}
        \, \text{and} \quad
        r_c=
        \begin{pmatrix}
            \log r_p \\ \log r_e \\ \log r_{eq}
            \end{pmatrix}
    \end{gathered}
    \label{eq:linear_system}
    \end{equation}

Each of the three constraints introduced before defines a plane in the 3D log-space of
configurations $h$.
Figure~\ref{fig:constraints} gives an illustration of these constraints.
Each plane on this figure constitutes a set of points where the associated contraint is zero (iso-constraint plane).
Parallel planes would be isoconstraint planes.
The determinant of this system $\det(A_{th})=-1$ is non-zero and therefore the system has a unique solution $y^\star$.
(of course this solution is unknown before doing hyperparameter search on $(\Lambda_p, \Lambda_e, b)$ because the values in $r_c$ are unknown).

\subsubsection{Comparison to experimental results}

Under this hypothesis it is natural to perform HPO by doing coordinate descent in the new system defined by the rows of $A_{th}$.
Unfortunately, the empirical geometry of evaluation performance landscape in the space of $h$ presented in ~\ref{fig:hp_omniglot} does not
match the directions found in $A_{th}$, and performing HPO in such a coordinate system also strongly underperforms using the directions empirically
found and described in the main part of the paper (equation~\ref{eq:matrix_a}). Out of this we conclude that the initial
separation hypothesis (according to which the hyperparameterization of the overall problem can be deduced from separately setting each subproblem
to its correct batch size / learning rate ratio) is false. In other words, the two sub-tasks of a DML or SSL loss are probably too
intertwined for them to be considered and modeled separately in terms of learning dynamics.

\end{document}